\documentclass[cameraready]{Interspeech}
\usepackage{pifont}
\usepackage{makecell}
\usepackage{subcaption}

\usepackage{comment}

\title{Queer inclusion in speech datasets:\\ An audit and taxonomy of practical tensions}

\author[affiliation={1,2}]{Brooklyn}{Sheppard}
\author[affiliation={2}]{Anaelia}{Ovalle}
\author[affiliation={2}]{Adina}{Williams}
\author[affiliation={2}]{Levent}{Sagun}

\address{
    $^1$ University of Calgary \\
    $^2$ Meta FAIR 
}

\email{brooklyn.sheppard1@ucalgary.ca}


\begin{document}

\maketitle

\begin{abstract}
    In this paper, we examine speech datasets for their inclusion of LGBTQIA+, or \textit{queer}, voices and provide a taxonomy of tensions to better understand why there is a lack of such voices in current speech technology datasets. Through an audit of six diverse speech datasets, we find that measurable queer representation is low (0–1.4\% of speakers) - insufficient for robust disparity measurement. We take this community as a case study to consider what challenges and tensions are associated with collecting speech data from marginalized communities. For comparison, we audit an additional two datasets from the speech sciences that were created by, for, and with the queer community. We note that many customs in speech dataset collection efforts in AI and speech technology research may conflict with values emphasized in participatory approaches with marginalized communities, and provide a taxonomy describing these tensions.
\end{abstract}

\section{Introduction}
\label{sec:intro}

Biases in speech and language models have often been partly attributed to lack of representation in the training data, with those underrepresented in the data being subject to disproportionately higher error rates on a variety of downstream tasks, including medical misdiagnoses, hiring decisions, and crime convictions  \cite{chinta2024ai, albaroudi2024comprehensive, hofmann2024ai, haimson2021disproportionate, dorn2024harmful, thiago2021fighting}. In the realm of speech models, creating diverse datasets is notoriously difficult given that, in contrast to text datasets, they cannot be created via highly controlled templates or synthetic data \cite{zhao2018gender, dorn2024harmful}, and instead require recruiting and recording speakers of a variety of different language communities. For annotations of sensitive attributes related to speaker identity, such as gender, disability status, or sexual orientation, the challenges in ethical data collection are amplified in speech given that participation depends on feeling safe to share these sensitive attributes related to their identity and that speech data is generally harder to anonymize than text, potentially exposing participants to unintended harms such as re-identification or voice misappropriation \cite{hutiri2024not, sigurgeirsson2024just}.

Sociolinguistic research has identified that one's voice is perceived as carrying an abundance of identity related characteristics, including age, race, gender, and sexuality \cite{mack2012influence, holliday2023siri, kushins2014sounding, smyth2003male, junger2013sex, leung2018voice, winkler2007influences}. Given the deeply rooted connection between voice and identity, we contend that the lack of voices from a given community in diverse speech datasets may contribute to biases towards these identities in speech technologies. Researchers in linguistics and speech sciences have studied not only the phonetic, prosodic, and lexical variations of queer speech \cite{smyth2002phonetics, shar2018expressing, gaudio1994sounding, rosales2019stylistics, gratton2016resisting}, but also the use of one's voices in building and portraying a distinct identity related to one's gender identity or sexual orientation \cite{podesva2007phonation, munson2007loose, zimman2018transgender, nuyen2023impact, hughes2024acoustic}. With one's voice being inherently linked to sensitive identity characteristics, collecting speech data representative of the queer community is particularly difficult given the harms that could arise from having labeled queer speech widely available, such as creating `gaydar' technologies \cite{mcara2024ai}; however, the lack of queer voices in speech technology datasets risks speech models systematically underperforming for speakers of this group.

Indeed, there have been several investigations into the social biases of speech models, such as wav2vec2.0, HuBERT, and Mockingjay \cite{baevski2020wav2vec, hsu2021hubert, liu2020mockingjay}. These investigations have demonstrated biases including associating the speech of certain groups with more positive valence over others, such as abled over disabled speakers and European Americans over African Americans speakers \cite{slaughter2023pre}. Crucially, they found that these associations persist beyond lexical content and correlate with speaker identity attributes. Additionally, these biases within the embedded representation of speech propagate to downstream tasks revealing negative emotion associations with disabled, male, and African American speech \cite{slaughter2023pre} and higher error rates on ASR tasks \cite{koenecke2020racial, mengesha2021don, feng2021quantifying, liu2022towards, riviere2021asr4real}. Following the trend in text-based bias research, recent work in speech model bias has attempted to use similar measures and develop speech fairness datasets to investigate implicit and explicit biases in speech-based models along dimensions of gender \cite{harris2024modeling, clifton2020100}, ethnicity \cite{coraal}, first language \cite{zhao2018l2arctic}, and dialect \cite{englishdialects, sanabria2023edinburgh}.


Several previous studies have noted that there is a significant lack of speech research relating to transgender and non-binary folks \cite{sanchez2024beyond, tronnierbias}. That is, many widely used speech datasets used to train and evaluate models on downstream tasks such as ASR, speaker verification, or speech synthesis provide only binary labels for speaker gender; this hinders the development of more diverse and realistic synthetic voices and risks speech models performing significantly worse for groups that are not represented within the gender binary. Indeed, in an effort to investigate gender biases in speech models beyond the binary notion of gender, \cite{attanasio-etal-2024-twists} used the Common Voice 16.0 corpus to measure performance differences of various speech models across speakers who identified as ``male'', ``female'', or ``other''. They found that across all models tested, while some models favoured male speakers and others favoured female speakers, all models revealed significant performance degradation for speakers who identified their gender as ``other''. Given the very limited sample size of speakers in this category (8 to 86 depending on the language), the authors note that this is merely a preliminary investigation and that future work would benefit from similar inquiries with a larger sample of speakers with various gender identities. This, however, is not possible if speech data from this community does not exist. 

Thus, taking the LGBTQIA+, or \textit{queer}, community as a case study, the goal of this work is to audit diverse speech datasets for their inclusion of queer voices, and through an analysis of dataset collection practices, we aim to identify why current speech data collection efforts tend to omit or underrepresent this community and provide a taxonomy of tensions between common AI data-collection practices and considerations for engaging with the queer community. Note that the present paper investigates the representation of queerness along the dimension of gender identity. This is an artifact of the fact that gender is a commonly provided annotation in speech datasets, while other queer identities are not. We believe, however, that the tensions identified apply to the queer community as a whole.

\section{Methodology}

\subsection{Dataset selection}
\label{sec:dat-sel}
We selected six English language speech technology datasets to audit based on three criteria:
(i) inclusion of diverse speaker backgrounds -- Edinburgh Accents (EdAcc) and English Dialects (Eng. Dia.) \cite{sanabria2023edinburgh, englishdialects};
(ii) previous use in assessing fairness in speech models -- L2-ARCTIC and Common Voice 22.0 (CV) \cite{zhao2018l2arctic, ardila2019common}; and
(iii) datasets specifically designed for speech fairness assessments -- Fairspeech and Casual Conversations V2 (CCV2) \cite{veliche2024towards, porgali2023casual}. That is, we audit open-source speech technology datasets that are likely to be used for evaluating performance disparities across speaker demographics. We audit an additional two datasets that were designed specifically for queer inclusion, namely, the Mid-Atlantic Gender Expansive Speech (MAGES) Corpus and a Palette of Transmasculine Voices (PTMV). The MAGES corpus consists of speech from 14 gender-expansive individuals, specifically those that don't identify within the gender binary \cite{mages-website}. The PTMV corpus consists of sentences spoken by 20 different masculine identifying participants \cite{dolquist2024clinical}. This includes cis-men, trans-men, trans non-binary folks, androgynous and more.

\subsection{Axes of auditing}

For each dataset, we ask the following questions regarding annotation practices, resulting distribution across gender labels, institutional affiliation, mention of an ethics review, recruitment methods, and data access: \textit{\textbf{Gender identity annotation options:}} Were gender labels self-provided? Were participants given pre-defined categories to choose from or free to provide their own label? \textit{\textbf{Distribution of gender labels:}} Of the datasets that include gender annotations beyond the binary, what percentage of the dataset consists of these speakers? \textit{\textbf{Institutional affiliation:}} Is the primary institution of the first author a for-profit corporation, not-for-profit organization or an academic institution? 
\textit{\textbf{Ethics review disclosure:}} Does the associated study explicitly describe an ethics or institutional review process?
\textit{\textbf{Recruitment methods:}} How were participants recruited? Was it through crowdsourcing techniques or engagement with specific underrepresented communities? If the latter, how did the researchers engage with the community? \textit{\textbf{Data access and usage:}} Is the dataset open access? Does the license allow for the dataset to be used for training and/or evaluating AI models?

\section{Comparison of datasets}

In the following sections, we outline the differences and similarities in these datasets regarding their methods for annotating gender categories, distribution across gender categories, their institutional affiliations and mention of an ethics review, their participant recruitment methods, and their data access and usage requirements. A summary of our comparison across datasets is provided in Table \ref{tab:compare}, with those created specifically with queer speakers in the last two lines. 

\begin{table*}[h!]
    \centering
    \begin{tabular}{c|ccc|ccccccc}
         Dataset  & \# Speakers & Institution & EthicsRD & Recruitment & Annotations & \makecell{Open \\ access?} & Train & Eval & \makecell{Queer\\Rep. (\%)} \\
         \hline
         CCV2 & $\sim$  5,567 & For-profit & \ding{55} & Paid speakers & 5 cat. & \ding{51} & \ding{55} & \ding{51} & 1.44 \\
         CV &  94,911 & Non-profit & \ding{55} & Crowdsourced & 4 cat. & \ding{51} & \ding{51} & \ding{51} & 0 \\
         EdAcc & 122 & Academic & \ding{51} & Crowdsourced & Self-provided & \ding{51} & \ding{51} & \ding{51} & 0.82 \\
         Eng. Dia. & 120 & For-profit& \ding{55} & Crowdsourced & Binary & \ding{51}  & \ding{51} & \ding{51} & 0\\
         Fairspeech &  593 & For-profit& \ding{55} & Paid speakers & Binary & \ding{51}& \ding{55} & \ding{51} & 0\\
         ARCTIC & 24 & Academic & \ding{55}  & Volunteers & Binary & \ding{51}& \ding{51} & \ding{51} & 0\\
         \hline
         MAGES & 14 & Academic & \ding{51} & \makecell{Word-of-mouth, \\ social media} & Self-provided & \ding{55}& \ding{51} & \ding{51} & 100 \\
         PTMV & 20 & Academic & \ding{51} & Social media  & Self-provided & \ding{51} & \ding{55} & \ding{55} & 45
    \end{tabular}
    \caption{Summary of speech datasets audited. Final two entries represent speech datasets created specifically for queer representation. EthicsRD (ethics review disclosed) indicates whether the associated study explicitly describes an ethics or institutional review process.}
    \label{tab:compare}
\end{table*}

\textbf{\textit{Gender annotation options:}} All datasets in our audit collect gender annotations via self-provided labels. The CV dataset allows for the mutually exclusive categories of self-provided gender annotations of ``male/masculine", ``female/feminine", ``non-binary", or ``transgender". The CCV2 dataset instead provides categories for ``cis-man", ``cis-woman", ``transgender man", ``transgender woman", and ``non-binary", but still requires gender identities to fit into pre-defined, mutually exclusive categories. In contrast, the MAGES and PTMV datasets allowed participants to freely define their gender identities in their own words. The only dataset for speech technology research that allowed for free text write in options for gender identity is that of the EdAcc dataset. Lastly, the Fairspeech, ARCTIC, and English Dialects datasets all contain strictly binary gender categories of male and female.

\begin{figure}[h!]
    \centering
    \begin{subfigure}{0.23\textwidth}        \includegraphics[width=1\linewidth]{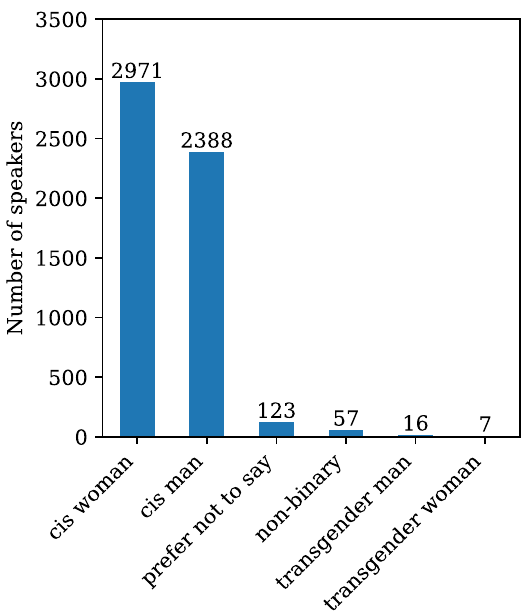}
    \caption{Casual Conversation V2}
    \label{fig:ccv2}
    \end{subfigure}
    \begin{subfigure}{0.23\textwidth}
        \includegraphics[width=1\linewidth]{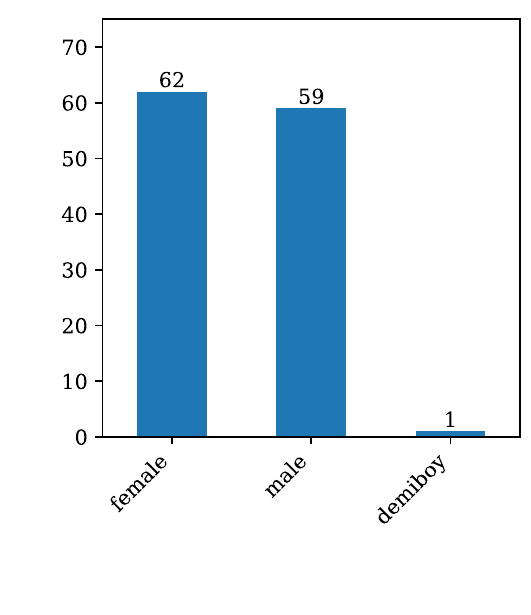}
    \caption{Edinburgh Accents}
    \label{fig:edacc}
    \end{subfigure}
    \caption{Distributions of speaker gender across two datasets.}
\end{figure}

\textbf{\textit{Distributions of gender labels:}} We next audit each dataset for queer inclusivity with an analysis of the distributions of gender categories. Given that the MAGES and PTMV datasets allowed individuals to self-express with no category restrictions and consist of predominantly queer speech, we do not provide distinct counts of gender identities for these datasets. All of the datasets do include gender annotations for each speaker, however, all but the CCV2 and EdAcc datasets include only binary gender labels. Interestingly, while both the CV and Fairspeech datasets had the option of identifying as outside the binary (e.g., ``non-binary", ``transgender") neither of these datasets contain any speech from participants that identify with these labels. In fact, as noted in the original Fairspeech paper, ``since we didn’t have a significant number of utterances from people who identified as non-binary, we chose to not include them, to not show skewed results" (\cite{liu2022towards}, p. 2). This suggests that while there were indeed some non-binary participants, the speech from these participants were excluded from the final dataset. On the other hand, the CV dataset simply has zero contributors that openly identify as either ``transgender" or ``non-binary". Figures \ref{fig:ccv2} and \ref{fig:edacc} demonstrate the counts of unique participants from each dataset that does contain data from some queer speakers (namely, CCV2 and EdAcc) by each possible gender category within the dataset.

\textbf{\textit{Institutional affiliations \& ethics review disclosure:}} 

Of the eight datasets considered, four come from academic institutions, three from large for-profit corporations, and one from a not-for-profit organization. Both queer-specific datasets were created by researchers at academic institutions. Additionally, only three datasets publicly disclosed an ethics or institutional review process prior to data collection: MAGES, PTMV, and EdAcc. We note, however, that lack of disclosure does not imply absence of internal ethical review, particularly for datasets produced within large industrial research organizations.

\textbf{\textit{Participant recruitment methods:}} The majority of datasets used crowdsourcing  methods, either through microwork platforms such as Fiverr or open calls to speakers, such as the CV dataset's online platform where anyone can contribute their voice to the dataset. Both the EdAcc and English Dialects datasets note that their initial data collection efforts stemmed from personal connections with the dataset creators. For example, the English Dialects dataset, created by Google, notes providing company-wide flyers asking for participation, including recruiting friends and family of current Google employees. On the other hand, the Fairspeech and CCV2, both created by Meta, do not provide details on exactly how they selected and recruited participants, but do indicate the speakers were paid participants. In contrast, both of the queer-specific speech datasets focused on community engagement via word-of-mouth and social media recruiting within the queer community.

\textbf{\textit{Data access and usage:}} All datasets considered here, with the exception of the MAGES corpus, are open access in that they are freely available online. That being said, the Terms of Use for several datasets restrict their use for certain purposes. Specifically, both the Fairspeech and CCV2 datasets may only be used to evaluate speech models, rather than train them. The CV, EdAcc, English Dialects, and ARCTIC corpora, however, are all available for both training and evaluating speech models, with the hope that they will be used to improve speech models on a variety of different speakers and languages. While the PTMV is open access and freely accessible, it states that its intended use case is specifically for clinical purposes, such as for speech pathologists to use as models of different expressions of masculinity for voice coaching those of the transmasculine community. The MAGES corpus, on the other hand, requires a meeting with the lead researcher to discuss your intentions with the data before it is released, in an effort to ``honor the vulnerability and safety of this community" \cite{mages-website}. Both queer speech datasets considered here were collected for the purpose of furthering research in the speech sciences, however, the MAGES corpus has also been used for modelling gender diverse voices in text-to-speech applications \cite{szekely2024inclusive}.

\section{Why is there a lack of queer voices in diverse speech datasets?}

From our audit of speech datasets and how they were collected, we provide a taxonomy of four tensions between what is commonly assumed or valued in the AI community versus the stated needs or preferences of the queer community: scaling versus inclusion, efficiency versus engagement, access versus autonomy, and static categories versus fluid identities. 


\subsection{Scaling versus inclusion}

The first tension we identify from our audit of diverse speech datasets is that of scaling versus inclusion. Scaling data collection efforts using crowdsourcing methods can be useful for collecting a large amount of data; many of the datasets discussed here focused on crowdsourcing efforts for their participant recruitment in an effort to collect a large amount of diverse speech. That is, the goal of these endeavours is to open data collection to all people in an effort be as representative as possible. This tension between scaling and inclusion is not inherent to statistical laws of scaling, but rather an issue of scaling in practice. That is, if enough data is sourced, it may theoretically be representative of all groups within a population, however, our audit shows that this may not necessarily be the case for all communities. A particularly interesting case of this tension is provided by the CV dataset. That is, the CV dataset is truly open and easy to contribute to - anyone with a microphone and an internet connection can donate their voice. This has allowed this effort to collect the largest amount of speech of all datasets discussed here. Despite this openness, however, the CV corpus has less queer representation than either the EdAcc or the CCV2 datasets, despite being 778 and 17 times their size in number of speakers. This tension thus suggests that, while scaling data collection efforts aim to be representative of all groups, our audit shows that, in practice, this method may still underrepresent small or stigmatized groups even with data from a large number of speakers. This finding is not entirely surprising as openly identifying as queer heightens the risk of discrimination and harm from surveillance, suggesting that smaller, community-focused data collection practices may alleviate this tension \cite{kalluri2025computer}.

\subsection{Efficiency versus engagement}

A related tension we identify from this audit is that of efficiency versus engagement. In addition to the goal of increasing the amount of data being collected, crowdsourcing efforts also have the benefit of being useful for collecting this data very efficiently. We argue that these methods may contradict the often slow process of engaging with affected community members to build trust, understand their needs, and develop AI systems and datasets collaboratively. 

For example, several datasets lack an explicit disclosure of an ethics or institutional review process in their papers; such processes can require researchers to reflect on the purpose and possible harms of their project prior to data collection.
While this is often the standard in the humanities, this is not the case in the AI community, despite many in recent years calling for this to change \cite{papakyriakopoulos2023augmented}. This process typically requires researchers to be explicit about their recruitment methods, benefits and risks to participants, and how potential harms might be mitigated. In the case of MAGES and PTMV datasets, participants were recruited via direct engagement with the affected community through social media and word-of-mouth initiatives by researchers who identify as part of the queer community, following ethics review. These datasets are of course significantly smaller than those provided by the AI community, reflecting the need for patience throughout the engagement process.

\subsection{Access versus autonomy}

Open access and open science is highly valued not only in the AI community, but the scientific community more generally, with efforts in open science having contributed significant advances in AI research. That being said, this value can conflict with the need for autonomy over one's own speech data, including how it is used and by whom. As  noted in Section \ref{sec:intro}, speech data is difficult to anonymize, potentially creating privacy risks for participants involved, but one's voice and speech may in many cases reflects personal identity related to gender identity or sexual orientation, and in turn can be of differing cultural significance to different individuals. All datasets considered here, with the exception of MAGES, are freely and openly available. 

Having openly available data of queer speech available to AI developers and the scientific community comes with potential risk to the community, such as contributing to the misappropriation of queer voices \cite{sigurgeirsson2024just}.
Indeed the MAGES corpus specifically notes that its closed access is in an effort to honor the vulnerability of this specific community. Interestingly, despite the PTMV corpus being specifically designed for, with and by members of the queer community, it resembles those lacking in queer representation in that it is open access. The key difference here is perhaps the way in which this data can be used. That is, the PTMV dataset was designed for clinical purposes in assisting gender diverse folks emulate a wide range of ``masculine" speech, while all other open-access datasets here are designed for speech technology research. In contrast to the clinical efforts of the PTMV dataset, the history of AI and technology includes instances where developments have adversely impacted the queer community. 

Some of these harms include using geolocation data to target and harm trans healthcare providers, predicting sexual orientation based on facial images, and the disproportionate censorship of LGBTQIA+ content through automated content-moderation systems and keyword-based filtering \cite{mcara2024ai}.

Additionally, qualitative interviews with trans and non-binary users of voice AI technologies (e.g., Siri and Alexa) indicate low levels of trust in developers’ ethics, business incentives, and labor practices. As described by participants, these concerns are often rooted in their perceptions of the broader sociotechnical contexts of development, specifically the structural power and inequalities within which these systems are produced \cite{rincon2021speaking}.

Scholars such as \cite{thatcher2016data} have proposed the concept of \textit{data colonialism} to describe asymmetries in data control, where academic or private institutions may inadvertently dispossess workers from the very data they create \cite{thatcher2016data}. This method of extracting data from communities directly conflicts with the concerns of queer folks related to the power and ideologies embedded within AI systems. 

\subsection{Static categories versus fluid identities}

All datasets audited included self-provided gender labels. While this practice is more desirable to third-party annotations on perceived gender \cite{papakyriakopoulos2023augmented}, in many cases, the options for gender labels are predefined and sparse. 

As demonstrated by PTMV participant self-descriptions combining trans, non-binary, and masculine identities, labels such as “trans” and “non-binary” are not necessarily mutually exclusive.
That is, many queer folks may identify as any combination of male, female, transgender, and non-binary. Thus, providing these pre-defined and mutually exclusive categories may discourage participation from those who aren't comfortable describing their gender identity within such strict confines. Additionally, the separation of the ``transgender" category from ``male" and ``female" within the CV dataset could create an implied distinction between ``transgender men" and ``men", and ``transgender women" and ``women". The practice of using discrete categories, while common in many AI and computer science applications for classification, can conflict with the fluid nature of identity. Indeed, the vast majority of algorithms and AI use-cases rely on some level of classification of the input. 

This practice, while sometimes useful and necessary, can sit uneasily with queer understandings of identity as dynamic, contextual, and not always reducible to distinct, mutually exclusive categories \cite{tomasev2021fairness,keyes2021you}.

This tension can be seen in the predominant practice of providing pre-defined gender labels in speech data collection as compared to the actual diversity of identities provided when queer folks are given the opportunity to self-express, as in the PTMV dataset. In many AI applications, categorization may be useful and necessary for some downstream task, however, the diversity in particular characteristics of one's voice across speakers and gender identities suggests that providing strict categories for gender identity is not necessary. In fact, the practice of categorizing queer voices may instead be harmful in opening up the possibility of automatic voice-based gender or sexuality identification; a practice that has the potential to harm queer individuals \cite{sigurgeirsson2024just}.

\section{Conclusion}

The present work has demonstrated the lack of queer representation in current diverse speech datasets, despite the fact that many of these datasets are theoretically open to all. Drawing on these findings and an analysis of the various dataset creation processes, we provide a taxonomy of tensions between customs in the AI community and the stated needs and preferences of the queer community.
We hope the present work brings attention to the lack of queer voices in otherwise diverse speech datasets and inspires future research on conducting participatory endeavors to engage with the queer community for the ethical development of speech and language datasets and technologies. Future work would benefit from directly engaging with queer speakers to understand these tensions through lived experience and how these tensions might differ across the various communities that exist under the queer umbrella. 

\section{Generative AI Use Disclosure}
The authors did not use any generative AI tools in any step of this research.

\bibliographystyle{IEEEtran}
\bibliography{mybib}

\end{document}